\documentclass[letterpaper, 10 pt, conference]{ieeeconf}
\IEEEoverridecommandlockouts
\usepackage{cite}
\usepackage{amsmath,amssymb,amsfonts}
\usepackage{algorithmic}
\usepackage{graphicx}
\usepackage{textcomp}

\usepackage{multirow}
\usepackage{adjustbox}
\usepackage{booktabs}
\usepackage{url}
\usepackage{xcolor}
\usepackage{color}
\usepackage{makecell}
\usepackage{ulem}
\usepackage[colorlinks=true,citecolor=blue,urlcolor=black]{hyperref}

\def\BibTeX{{\rm B\kern-.05em{\sc i\kern-.025em b}\kern-.08em
    T\kern-.1667em\lower.7ex\hbox{E}\kern-.125emX}}
\begin{document}
\title{\LARGE \bf SurgFusion-Net: Diversified Adaptive Multimodal Fusion Network for Surgical Skill Assessment}
\author{Runlong He, Freweini M. Tesfai, Matthew W. E. Boal, Nazir Sirajudeen, Dimitrios Anastasiou, Jialang Xu, \\ Mobarak I. Hoque, Philip J. Edwards, John D. Kelly, Ashwin Sridhar, Abdolrahim Kadkhodamohammadi, \\ Dhivya Chandrasekaran, Matthew J. Clarkson, Danail Stoyanov, \IEEEmembership{Fellow, IEEE}, Nader Francis, and \\ Evangelos B. Mazomenos, \IEEEmembership{Member, IEEE}
\thanks{This work was supported in whole, or in part, by Medtronic Digital Surgery, the EPSRC [EP/Z534754/1, EP/W00805X/1, EP/T517793/1, EP/S021930/1, UKRI145], the DSIT and the RAEng under the Chair in Emerging Technologies. For the purpose of open access, the authors have applied a CC BY public copyright licence to any author accepted manuscript version arising from this submission. (\textit{Corresponding authors: R. He, E. B. Mazomenos)}}
\thanks{R. He, N. Sirajudeen, D. Anastasiou, J. Xu, M. J. Clarkson, and E. B. Mazomenos are with the UCL Hawkes Institute and the Department of Medical Physics \& Biomedical Engineering, UCL, UK. (E-mail: \{runlong.he.23, nazir.sirajudeen.20, dimitrios.anastasiou.21, jialang.xu.22, m.clarkson, e.mazomenos\}@ucl.ac.uk).}
\thanks{P. J. Edwards, and D. Stoyanov are with the UCL Hawkes Institute and the Department of Computer Science, UCL, UK. (E-mail: \{eddie.edwards,    danail.stoyanov\}@ucl.ac.uk).}
\thanks{M. I. Hoque is with the UCL Hawkes Institute and the Division of Informatics, Imaging \& Data Sciences, The University of Manchester, UK (E-mail: mobarak.hoque@manchester.ac.uk).}
\thanks{J. D. Kelly, A. Sridhar and D. Chandrasekaran are with the UCL Hospitals NHS Foundation Trust, UK. (E-mail: \{justin.collins; j.d.kelly, d.chandrasekaran\}@ucl.ac.uk, ashwin.sridhar@nhs.net).}
\thanks{F. M. Tesfai, M. W. E. Boal, and N. Francis are with the Griffin Institute, Northwick Park and St Mark’s Hospital, UK. (E-mail: \{f.tesfai, m.boal, n.francis\}@griffininstitute.org.uk).}
\thanks{Abdolrahim Kadkhodamohammadi is with Medtronic Digital Surgery, London, UK (E-mail: rahim.mohammadi@medtronic.com).}
}

\maketitle

\begin{abstract}
Robotic-assisted surgery (RAS) is established in clinical practice, and automated surgical skill assessment utilizing multimodal data offers transformative potential for surgical analytics and education. 
However, developing effective multimodal methods remains challenging due to the task complexity, limited annotated datasets and insufficient techniques for cross-modal information fusion.
Existing state-of-the-art relies exclusively on RGB video and only applies on dry-lab settings, failing to address the significant domain gap between controlled simulation and real clinical cases, where the surgical environment together with camera and tissue motion introduce substantial complexities.
This work introduces SurgFusion-Net and Divergence Regulated Attention (DRA), an innovative fusion strategy for multimodal surgical skill assessment. We contribute two first-of-their-kind clinical datasets: the RAH-skill dataset containing 279,691 RGB frames from 37 videos of Robot-assisted Hysterectomy (RAH), and the RARP-skill dataset containing 70,661 RGB frames from 33 videos of Robot-Assisted Radical Prostatectomy (RARP). Both datasets include M-GEARS skill annotations, corresponding optical flow and tool segmentation masks. DRA incorporates adaptive dual attention and diversity-promoting multi-head attention to fuse multimodal information, from three modalities, based on surgical context, enhancing assessment accuracy and reliability.
Validated on the JIGSAWS benchmark, RAH-skill, and RARP-skill datasets, our approach outperforms recent baselines with SCC improvements of 0.02 in LOSO, 0.04 in LOUO across JIGSAWS tasks, and 0.0538 and 0.0493 gains on RAH-skill and RARP-skill, respectively.
Our source code and dataset is available at~\url{https://github.com/HRL-Mike/SurgFusion-Net}.
\end{abstract}

\section{INTRODUCTION}
\label{sec:introduction}
The rapid development and clinical application of robot-assisted surgery (RAS) has revolutionized modern surgical healthcare, with advantages in precision and safety, improving patient outcomes~\cite{moglia2021systematic}. However, the complexity of RAS requires surgeons to undergo rigorous professional training to achieve clinical competency. Traditional skill assessment relies on expert review by senior surgeons and manual marking of Liker-point rating scales such as the Global Evaluative Assessment of Robotic Skills (GEARS)~\cite{sanchez2016robotic}. Broad deployment of such tools remains impractical due to restrictive time requirements compounded by inter-rater variability and limited standardization~\cite{boal2024evaluation}. With the proliferation of RAS technology and growing surgical demands, establishing automated and objective surgical skill assessment systems has become an urgent need for supporting surgical training and ensuring optimal operational performance~\cite{walshaw2025essential}. 

Immediately-accessible, objective performance evaluation can substantially benefit surgical education, streamline experts’ roles and resources and ultimately accelerate the learning curve of trainees~\cite{ackermann2023factors}. In clinical cases, it can produce key insights for downstream tasks, such as intelligent navigation for increased safety~\cite{quarez2025recap}. However, realizing automated RAS skill assessment is an intricate task that requires contextualization and quality evaluation of human (i.e. surgeon) actions. Prior work is limited to dry-lab training environments, primarily the JIGSAWS dataset~\cite{ahmidi2017dataset, van2022gesture}, while the applicability of developed methods is yet to be demonstrated in real clinical cases. The state-of-the-art (SOTA) includes video-based, single-modal approaches~\cite{anastasiou2023keep, li2022surgical}, which although effective on controlled training tasks, their adoption in real cases is compromised by the complex visual characteristics and dynamic nature of intraoperative videos~\cite{ayobi2025pixel}. At the same time, current multimodal fusion methods for skill assessment~\cite{zeng2024multimodal} rely on unidirectional attention flow and attention head homogenization, limiting their capacity to model complex, inter-modal relationships beyond simple similarity-based patterns. These challenges motivate the development of innovative multimodal fusion approaches and new clinical datasets.

This work presents SurgFusion-Net, a novel method for RAS skill assessment integrating optical flow and tool segmentation as additional inputs to standard RGB. Our multimodal approach characterizes the instruments' spatial-temporal dynamics while filtering background noise. Encoded features are fused via Divergence Regulated Attention (DRA), an innovative fusion technique, which adaptively modulates positive and negative attention to learn consistent and complementary representations across the three modalities, while incorporating a lightweight divergence constraint to promote multi-head diversity. Our work also introduces the first clinical datasets for RAS skill assessment of suturing tasks during the vaginal vault closure phase in Robot-assisted Hysterectomy (RAH) and the dorsal venous complex ligation in Robot-assisted Radical Prostatectomy (RARP). The datasets include RGB video, optical flow maps, tool segmentation masks and annotations using the Modifiable Global Evaluative Assessment of Robotic Skills (M-GEARS) tool~\cite{boal2024development}. SurgFusion-Net achieves superior estimation of RAS skill scores with excellent consistency and improvements of 0.02-0.054 in Spearman's Correlation Coefficient (SCC), outperforming current SOTA methods on both JIGSAWS and our two clinical datasets. 
Our main contributions are summarized as follows:
\begin{enumerate}
    \item[--] We build RAH-skill and RARP-skill, two first-of-their-kind multimodal clinical datasets for RAS skill assessment containing RGB videos, optical flow maps, tool segmentation masks and M-GEARS annotations. RAH-skill has 279,691 frames and RARP-skill 70,661 frames from 37 RAH and 33 RARP cases respectively.

    \item[--] We propose Divergence Regulated Attention (DRA), a multimodal fusion technique tailored for RAS skill assessment, that adaptively balances positive and negative attention to capture cross-modal features while ensuring multi-head diversity through divergence constraints.
    
    \item[--] We propose SurgFusion-Net, a multimodal network that progressively fuses RGB features and, for the first time, optical flow and segmentation features through cross-stage and dynamic fusion blocks with DRA, to capture surgical scene representations for RAS skill assessment. 
    
    \item[--] We evaluate SurgFusion-Net on the JIGSAWS benchmark and our two clinical datasets using both SCC and mean absolute error (MAE). Our approach outperforms the SOTA by 0.04 on JIGSAWS across-tasks in the LOUO setup, by 0.0538 on RAH-skill, and by 0.0493 on RARP-skill. MAE is reduced by 0.36, 0.1463, and 0.0614 on the three datasets. Ablation studies validate the effectiveness of DRA and SurgFusion-Net.
\end{enumerate}

\section{RELATED WORK}
\subsection{Action Quality Assessment}
Action quality assessment (AQA) is a fine-grained action understanding task that aims to assess the quality of performed actions, providing objective and automated assessments across domains such as sports analysis, medical care, and professional skill evaluation. Existing AQA works are categorized based on video length: methods for short videos (averaging several seconds) and methods for long videos (averaging several minutes). 
Significant progress has been made in short video AQA, with examples including spatio-temporal contextual networks~\cite{wang2021tsa} and temporal parsing transformers for fine-grained representations~\cite{bai2022action}.
For long videos, a key challenge is that segmentation into uniform clips leads to loss of contextual information. Recent works~\cite{gao2023automatic,gedamu2024self} address this by shifting towards video-level representation learning, which aggregates clip-level features into comprehensive video representations.
While early AQA methods primarily relied on visual information, multimodal approaches have emerged as a promising trend. Majeedi et al.~\cite{majeedi2024rica2} proposed a deep probabilistic model combining textual descriptions with video for estimating judges' scores in Olympic diving. Zhang et al. combined video and audio-transcribed text to produce action narratives with scores~\cite{zhang2024narrative}. Zeng et al. integrate audio with video and optical flow for AQA in figure skating and gymnastics~\cite{zeng2024multimodal}, showing multimodal integration.
Our work is unique in multimodal AQA as it introduces DRA, that adaptively learns consistent and complementary patterns across modalities, surpassing unidirectional attention flow approaches.

\begin{figure*}[!ht]
    \centering
    \includegraphics[width=0.95\textwidth]
    {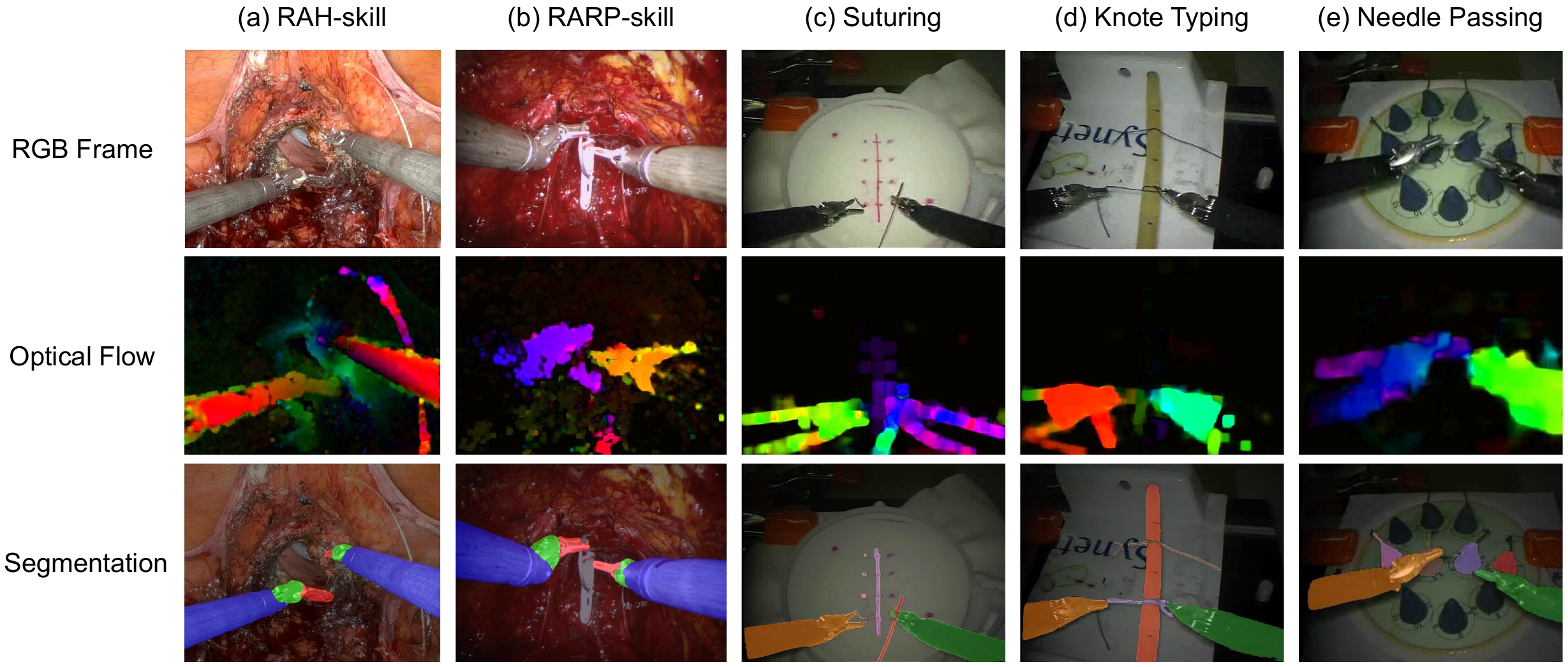}
    \caption{Examples from the multimodal RAH-skill and RARP-skill datasets, along with the enhanced JIGSAWS dataset. JIGSAWS comprises three tasks: Suturing, Knot Tying, and Needle Passing. All datasets contain RGB frames, optical flow maps, and segmentation masks. In optical flow maps, black background indicates no motion, brightness represents motion speed, and colors indicate motion directions. RAH-skill and RARP-skill include tool segmentation masks, while JIGSAWS contains both tool and reference object masks.}
    \label{fig:datasets}
\end{figure*}

\subsection{Skill Assessment in RAS}
Most prior work is based on the JIGSAWS benchmark, which contains synchronized kinematic and video data of basic training tasks (needle passing, knot tying, suturing) with the da Vinci surgical system (Intuitive Surgical Inc, USA) alongside gesture and Global Rating Scale (GRS) scores annotations~\cite{ahmidi2017dataset}.
Early studies concentrated on kinematics feature engineering~\cite{zia2018automated} and temporal convolutional networks~\cite{fawaz2019accurate} to regress GRS scores from robotic instrument trajectories.
The SOTA on JIGSAWS mostly utilizes the RGB video alone, employing spatial-temporal encoders to learn discriminative representations. Wang et al.~\cite{wang2020towards} capture temporal differences using surgical gesture information. Li et al.~\cite{li2022surgical} model tool-tissue interactions by clustering semantic features and using bidirectional LSTMs for score regression. Anastasiou et al.~\cite{anastasiou2023keep} employ contrastive learning to predict GRS scores through similarity modeling. A multipath approach combining video, kinematics and gesture sequencing is presented in~\cite{liu2021towards}, also tested on a private laparoscopic cholecystecomy dataset. Results from these methods show considerable variability across the three JIGSAWS tasks and cross-validation setups. In our view, this occurs because features from RGB alone inherently contain noise and interference from irrelevant background areas, while single-point (tool tip) kinematics cannot capture adequate information for skill estimation. These limitations will be even more pronounced in dynamic clinical cases. Although spearheading research on RAS skill assessment, JIGSAWS cannot replicate the visual complexities of real surgical environments, particularly frequent camera adjustments and background tissue motion. To the best of our knowledge the only open-source clinical dataset with skill annotations is from laparoscopic cholecystectomies (HeiChole), but without extended benchmarking~\cite{wagner2023comparative}. 
Our work represents the first attempt to explore multimodal visual information in RAS skill assessment, using optical flow and tool segmentation to mitigate background interference and motion artifacts. This enables SurgFusion-Net to achieve both accurate and consistent estimation of skill scores in JIGSAWS and crucially in two new RAS clinical datasets.

\section{METHODOLOGY}
\subsection{Datasets and Annotation}
\subsubsection{RAH-skill dataset}
The RAH-skill Dataset consists of 37 videos of the vaginal vault closure phase from RAH cases collected at University College London Hospital and Yeovil Hospital under ethical approval and patient consent (IRAS ID: 309024).
The surgeries were performed by 11 surgeons across three expertise levels: beginners (n=3, $<$10 cases), intermediate (n=5, 10-59 cases), and experts (n=3, $>$60 cases). Videos were recorded at 30 Hz and stored in “.mp4" format.
The dataset comprises 9 hours 35 minutes of footage, with videos averaging 15 minutes 34 seconds (ranging from 6:26 to 27:46). Each video was segmented into three phases: right angle securement, middle section closure, and left angle securement. Non-suturing frames were removed, and each phase was scored using the M-GEARS tool~\cite{boal2024development}.
All videos were blindly double-annotated by two trained surgeons with 90\% or higher inter-rater agreement on pre-validated videos. Final scores represent the mean of the two raters, yielding 110 M-GEARS scores ranging from 14 to 30 across all phases, with strong inter-rater reliability (Spearman rho: $\rho$ = 0.756, 95\% CI [0.369, 0.908], p = 0.002).

\subsubsection{RARP-skill dataset}
The RARP-skill dataset contains 33 videos, a subset of the open-source SAR-RARP50 dataset, of dorsal venous complex ligation from RARP collected at University College Hospital (Westmoreland Street, London, UK). The surgeries were performed by 8 surgeons across three expertise levels, including consultants, senior and junior registrars~\cite{psychogyios2023sar}. Videos were recorded at 60 Hz and stored in “.avi” format. Based on video quality, 33 videos were selected for annotation, comprising 2 hours 20 minutes of footage with an average length of 4 minutes 16 seconds (ranging from 1:40 to 10:32). A total of 20 videos were blindly double-annotated by a trained and a resident surgeon, with the rest annotated by the trained surgeon and reviewed by the resident surgeon. Inter-rater agreement (85\%) was established on 10 suturing videos from JIGSAWS and discrepancies were resolved through joint review to reach consensus. This yielded 33 M-GEARS scores ranging from 18 to 28. Strong inter-rater reliability was confirmed on the 20 videos for both absolute M-GEARS scores (Spearman rho: $\rho$ = 0.7026, 95\% CI [0.3774, 0.8735], p $<$ 0.001) and score ranking (Spearman rho: $\rho$ = 0.7281, 95\% CI [0.4214, 0.8854], p $<$ 0.001)~\cite{sirajudeen24deep}.

\subsubsection{Data Processing}
RGB frames were extracted from all videos at 10 FPS, yielding 279,691 frames for RAH-skill and 70,661 frames for RARP-skill, with averages of 7,559 and 2,141 frames per video, respectively.
Optical flow was computed using Farnebäck algorithm between consecutive frames, with x and y components scaled to [0, 255], yielding 559,162 optical flow maps for RAH-skill and 141,256 for RARP-skill. Tool segmentation masks were extracted using a personalized federated learning model~\cite{xu2025personalizing} in a zero-shot manner, obtaining 279,691 masks for RAH-skill and  70,661 for RARP-skill. This model, trained on EndoVis 2017/2018~\cite{allan20172017,allan20182018} and SAR-RARP50~\cite{psychogyios2023sar} datasets, achieved 85.18\% Dice score across surgical instruments.
Fig.~\ref{fig:datasets} shows examples from the RAH-skill (a) and RARP-skill (b) datasets.

\begin{figure*}[!ht]
    \centering
    \includegraphics[width=0.99\textwidth]
    {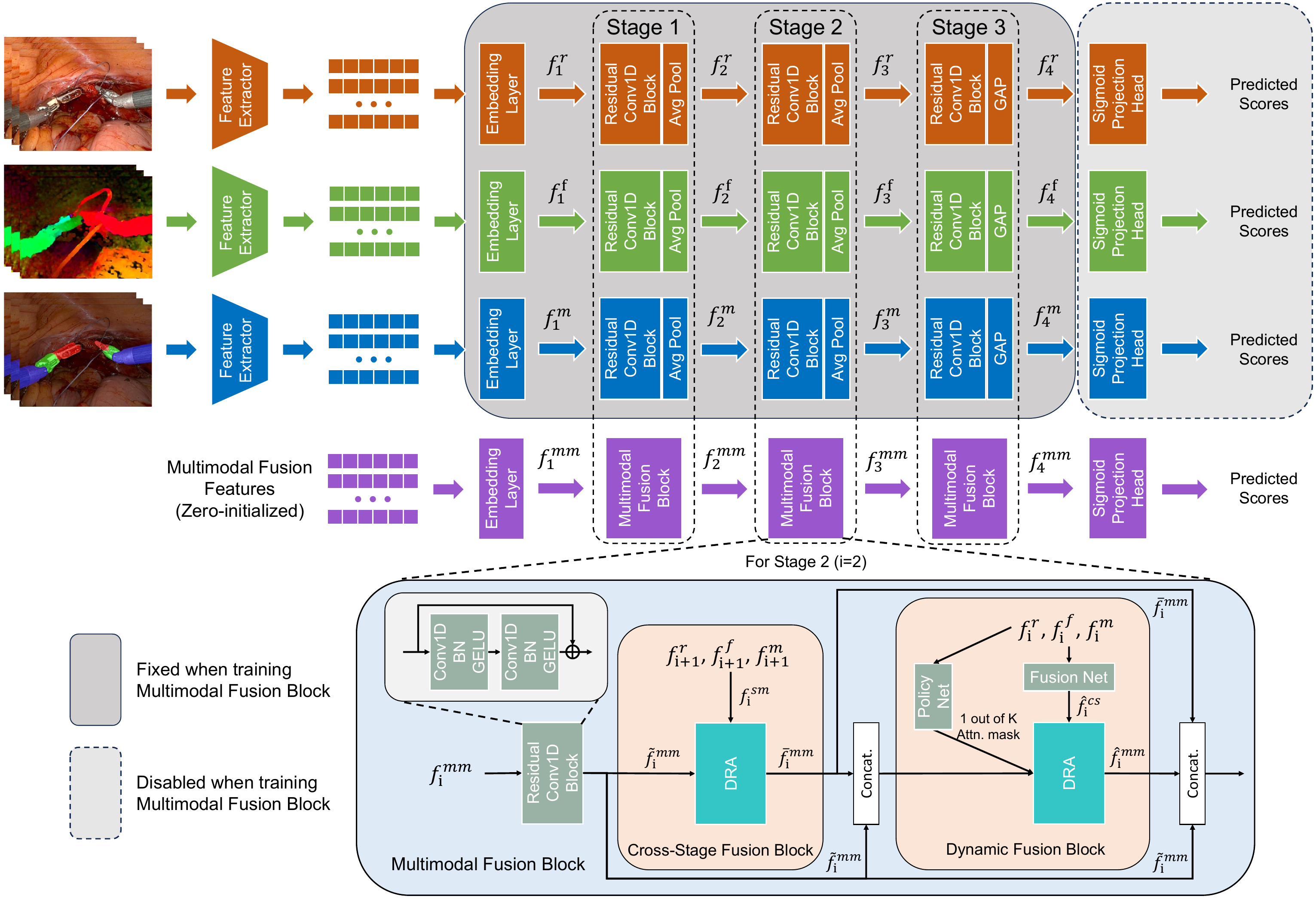}
    \caption{SurgFusion-Net: the network consists of three unimodal branches and a multimodal fusion branch. The architecture progressively fuses unimodal features to construct comprehensive multimodal representations for robotic surgical skill assessment.}
    \label{fig:model_archi}
\end{figure*}

\subsection{Proposed Method}
\subsubsection{Problem Formulation}
We formulate surgical skill assessment as a regression problem, where the model observes a video containing a series of surgical actions and predicts a non-negative value as the video skill score. We normalize the labels to $[0, 0.5]$ for stable training.
 
\subsubsection{SurgFusion-Net}

\textit{Model Structure Overview:}
To leverage multimodal information for RAS skill assessment, we propose SurgFusion-Net, a novel multimodal fusion network. SurgFusion-Net independently learns information from different modalities and progressively constructs fused multimodal representations from unimodal features.
As illustrated in Fig.~\ref{fig:model_archi}, SurgFusion-Net comprises four branches: RGB, optical flow, segmentation mask, and multimodal fusion. We first train the three unimodal branches to learn modality-specific features, then freeze their backbone networks and disable regression heads to extract features for fusion.
The multimodal fusion branch utilizes cross-stage and dynamic fusion blocks to update zero-initialized fusion features with extracted unimodal features, ultimately predicting surgical video scores.

\textit{Unimodal Branch:}
To extract unimodal information, all three branches are pretrained and fixed during fusion training. Each branch consists of three convolutional stages and a regression layer, with each stage containing a residual convolutional block and pooling.
For the $i$-th stage, we denote inputs and outputs as $\{f_i^r, f_{i+1}^r\}$ for RGB, $\{f_i^f, f_{i+1}^f\}$ for optical flow, and $\{f_i^m, f_{i+1}^m\}$ for segmentation mask. Each residual block follows~\cite{he2016deep} but uses 1D convolutions. The block processes the input through two consecutive stages, each consisting of 1D convolution, batch normalization and GELU activation, then adds a residual connection:
\begin{equation}
\tilde{f}_i^r = f_i^r + \text{Conv}(\text{Conv}(f_i^r))
\end{equation}

Subsequently, average pooling is applied to reduce the temporal dimension:
\begin{equation}
f_{i+1}^r = \text{AvgPool}(\tilde{f}_i^r)
\end{equation}

where $\tilde{f}_i^r$ is the residual block output. The final regression layer uses a fully-connected layer with dropout and sigmoid activation to predict skill scores.

\textit{Multimodal Fusion Branch:}
After obtaining unimodal features, a multimodal fusion branch progressively aggregates features from all modalities. This branch follows a similar architecture to unimodal branches but employs multimodal fusion blocks for cross-modal integration. Learnable fusion features are initialized to zeros and updated by these blocks. To enable information transfer from unimodal branches to multimodal branch, we propose two novel modules: cross-stage fusion block and dynamic fusion block.

\textbf{1). Cross-Stage Fusion Block (CSFB):}
The Cross-Stage Fusion Block utilizes stage $i$ outputs $\{f_{i+1}^r, f_{i+1}^f, f_{i+1}^m\}$ from unimodal branches to refine fusion features at the same stage. CSFB integrates a DRA block that dynamically captures cross-modal dependencies while promoting attention head diversification.
Inspired by~\cite{vaswani2017attention}, DRA is implemented through cross-attention, with details in Section~\ref{sec:dra}. The query is obtained by projecting the multimodal fusion features after the residual block $\tilde{f}_{i}^{mm}$, while keys and values are obtained by projecting the concatenated stage $i$ unimodal outputs:
\begin{equation}
Q_{i} = \mathbf{W}^q \tilde{f}_{i}^{mm}, \quad K_{i} = \mathbf{W}^k f_{i}^{sm}, \quad V_{i} = \mathbf{W}^v f_{i}^{sm}
\end{equation}
\begin{equation}
f_{i}^{sm} = \text{Concat}(f_{i+1}^r, f_{i+1}^f, f_{i+1}^m)
\end{equation}
where $\mathbf{W}^q$, $\mathbf{W}^k$, and $\mathbf{W}^v$ are learnable projection matrices. All features $f_{i+1}^r$, $f_{i+1}^f$, $f_{i+1}^m$ and $\tilde{f}_{i}^{mm}$ share the same dimension $\mathbb{R}^{T_i \times d}$, where $T_i$ is the sequence length at stage $i$. The multimodal fusion features are then processed through DRA:
\begin{equation}
\bar{f}_{i}^{mm} = \text{DRA}(Q_{i}, K_{i}, V_{i})
\end{equation}

This design leverages the specialized unimodal representations ${f_{i+1}^r, f_{i+1}^f, f_{i+1}^m}$ from the $i$-th stage, rather than the output features from the previous stage, to enhance cross-modal information exchange and fusion.

\textbf{2). Dynamic Fusion Block (DFB):}
Since different surgical actions and scenes exhibit varying complexity and require different fusion strategies, we employ a Dynamic Fusion Block (DFB) to learn flexible multimodal fusion strategies. Following~\cite{zeng2024multimodal}, the DFB incorporates K FusionNets to explore diverse fusion strategies and a PolicyNet to adaptively select which strategies to enable. We replace the Cross-modal Feature Decoder in~\cite{zeng2024multimodal} with our DRA for enhanced decoding capabilities. The PolicyNet generates an optimized attention mask $M_{i}$ at the $i$-th stage to guide the DRA decoding process.
The query of DRA within DFB is formed by concatenating the residual block output $\tilde{f}{i}^{mm}$ and the CSFD output $\bar{f}{i}^{mm}$:
\begin{equation}
\hat{Q}_i = \hat{\mathbf{W}}^q \cdot \text{Concat}(\tilde{f}_{i}^{mm}, \bar{f}_{i}^{mm})
\end{equation}

The keys and values are formed by concatenating the cross-modal features $f_{i,k}^{cs}$ from K FusionNets:
\begin{equation}
\hat{K}_{i} = \hat{\mathbf{W}}^k \cdot \bar{f}_{i}^{cs}, \quad \hat{V}_{i} = \hat{\mathbf{W}}^v \cdot \bar{f}_{i}^{cs}
\end{equation}
\begin{equation}
\bar{f}_{i}^{cs} = \text{Concat}(f_{i,1}^{cs}, f_{i,2}^{cs}, \ldots, f_{i,K}^{cs})
\end{equation}
\begin{equation}
f_{i,k}^{cs} = \alpha_{i,k}^{r} f_{i}^{r} + \alpha_{i,k}^{f} f_{i}^{f} + \alpha_{i,k}^{m} f_{i}^{m}
\end{equation}
where $\hat{\mathbf{W}}^q$, $\hat{\mathbf{W}}^k$, and $\hat{\mathbf{W}}^v$ are learnable projection matrices, $\alpha_{i, k}^{r}$, $\alpha_{i,k}^{f}$ and $\alpha_{i,k}^{m}$ are fusion weights from the $k$-th FusionNet, and $f_{i}^{r}$, $f_{i}^{f}$, $f_{i}^{m}$ represent the input RGB, optical flow, and segmentation features at stage $i$, respectively.
The multimodal fusion features from DFB are then computed with DRA as:
\begin{equation}
\hat{f}_{i}^{mm} = \text{DRA}(\hat{Q}_i, \hat{K}_i, \hat{V}_i, M_i)
\end{equation}

The DFB employs a 1-out-of-K attention mask from the PolicyNet to guide DRA decoding, enhancing its ability to adaptively leverage diversified attention for complex surgical scene understanding.
\begin{figure}[!ht]
    \centering
    \includegraphics[width=\columnwidth]{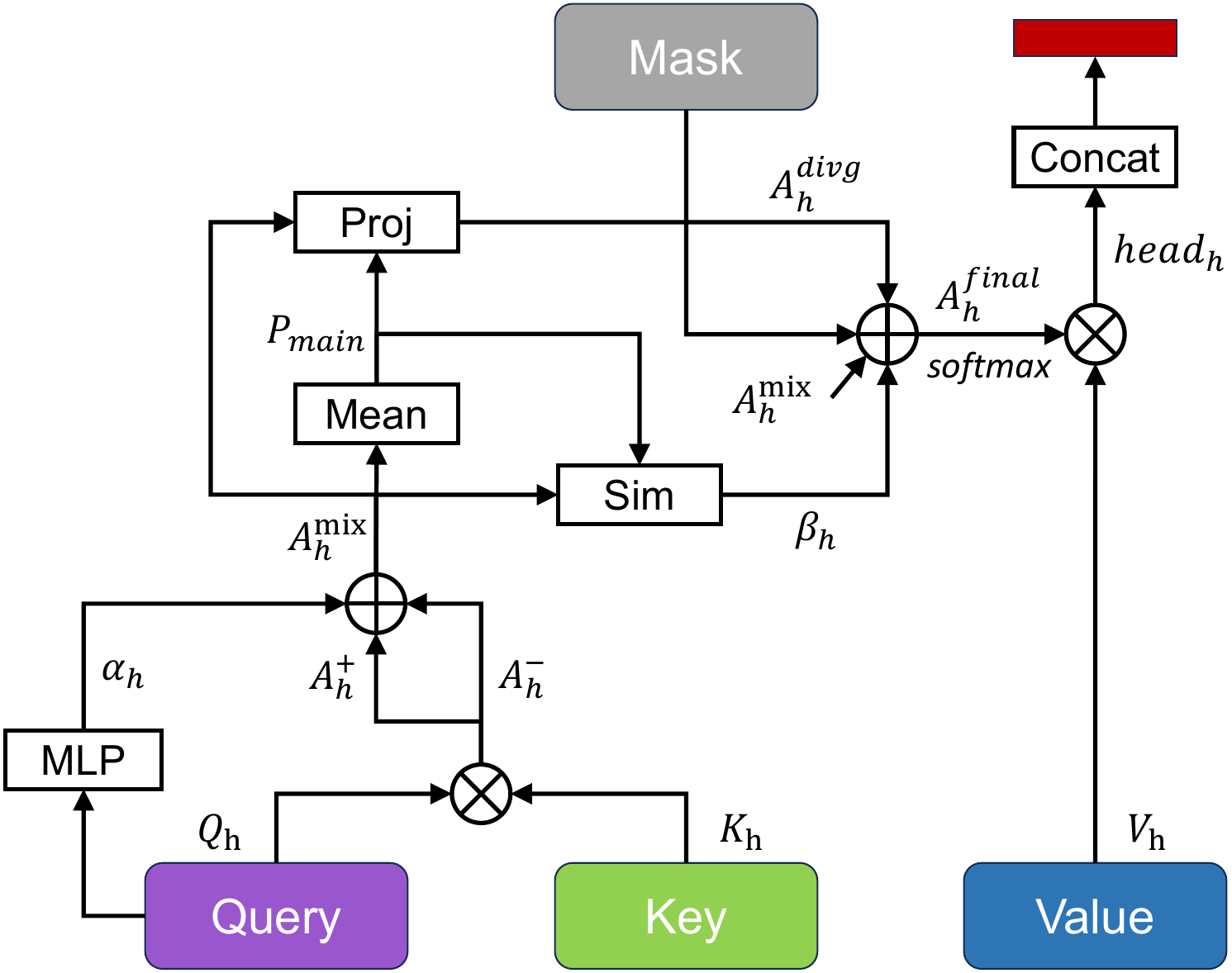}
    \caption{Computation flow of Divergence Regulated Attention. $Q_h$, $K_h$, $V_h$ are query, key, value representations for each head. $\otimes$: matrix multiplication; $\oplus$: element-wise addition; Proj: projection; Sim: similarity.}
    \label{fig:DRA}
\end{figure}
\subsubsection{Divergence Regulated Attention} \label{sec:dra}
We propose DRA that combines vanilla attention~\cite{vaswani2017attention} and complementary attention~\cite{zeng2024multimodal} with adaptive weight modulation and encourages diversity across attention heads. The DRA computation flow is shown in Fig.~\ref{fig:DRA}.

Given the input query $Q \in \mathbb{R}^{b \times s_q \times d}$, keys $K \in \mathbb{R}^{b \times s_k \times d}$, and values $V \in \mathbb{R}^{b \times s_v \times d}$, where $b$ is the batch size, $s_q$, $s_k$, $s_v$ are the sequence lengths, and $d$ is the feature dimension, DRA employs three separate linear projection layers to generate query, key, and value representations. The projected features are then divided across $H$ attention heads, where each head $h$ operates on dimension $d_h = d / H$. For each attention head $h$, the adaptive attention mechanism computes both positive and complementary attention scores:
\begin{equation} \label{eq:A_pls}
\mathcal{A}_h^+ = \frac{Q_h K_h^T}{\sqrt{d_h}}, \quad \mathcal{A}_h^- = -\mathcal{A}_h^+
\end{equation}

To adaptively balance these two attention patterns, we introduce a content-aware mixing factor for each head:
\begin{equation}
\alpha_h = \sigma(\text{MLP}(\text{AvgPool}(Q_h)))
\end{equation}
where $\sigma$ is the sigmoid function. This head-specific mixing factor enables each attention head to automatically adjust its attention strategy based on the input content, providing flexible and context-sensitive attention mechanisms.
The mixed attention $\mathcal{A}_h^{mix}$ for each head is then computed as:
\begin{equation} \label{eq:A_mix}
\mathcal{A}_h^{mix} = \alpha_h \cdot \mathcal{A}_h^+ + (1-\alpha_h) \cdot \mathcal{A}_h^-
\end{equation}

To promote diversity across attention heads, we apply a lightweight divergence constraint based on principal components. The main attention pattern $\mathcal{P}_{main}$ is computed as:
\begin{equation} \label{eq:main_pattern}
\mathcal{P}_{main} = \text{Mean}(\mathcal{A}_h^{mix}, \text{dim}=\text{head})
\end{equation}

The diversity factor $\beta_h$ encourages each head to deviate from the dominant pattern:
\begin{equation} \label{eq:beta_factor}
\beta_h = \sigma\left(1 - \frac{\langle \mathcal{A}_h^{mix}, \mathcal{P}_{main} \rangle}{||\mathcal{P}_{main}||^2 + \epsilon}\right)
\end{equation}
where $\epsilon = 10^{-8}$ ensures numerical stability. The projection coefficient $\left( \langle \mathcal{A}_h^{mix}, \mathcal{P}_{main} \rangle \right) / \left( ||\mathcal{P}_{main}||^2 + \epsilon \right)$ quantifies how aligned each attention head is with the dominant pattern. 
We define the divergence component $\mathcal{A}_h^{divg}$ that captures each head's unique deviation from the main pattern:
\begin{equation} \label{eq:A_divg}
\mathcal{A}_h^{divg} = \mathcal{A}_h^{mix} - \frac{\langle \mathcal{A}_h^{mix}, \mathcal{P}_{main} \rangle}{||\mathcal{P}_{main}||^2 + \epsilon} \cdot \mathcal{P}_{main}
\end{equation}

The final diversified attention $\mathcal{A}_h^{final}$ is then constructed by adding $\mathcal{A}_h^{divg}$, scaled by the diversity factor $\beta_h$:
\begin{equation} \label{eq:A_final}
\mathcal{A}_h^{final} = \mathcal{A}_h^{mix} + \beta_h \cdot \mathcal{A}_h^{divg}
\end{equation}

Based on Eq.~\ref{eq:A_mix} to~\ref{eq:A_final}, when $\mathcal{A}_h^{mix}$ is similar to $\mathcal{P}_{main}$, smaller values of $\beta_h$ and $\mathcal{A}_h^{divg}$ provide modest diversity promotion, while differences between attention heads amplify the divergence component to promote stronger diversification.
The output for each head is then obtained through softmax-weighted aggregation:
\begin{equation}
\text{head}_h = \text{softmax}(\mathcal{A}_h^{final}) \cdot V_h
\end{equation}

The final output is obtained by concatenating all heads: $\text{MultiHead} = \text{Concat}(\text{head}_1, \text{head}_2, \ldots, \text{head}_H)$.
This mechanism enables the model to adaptively balance consistent and complementary information based on surgical context while encouraging diverse attention patterns across different heads.

\subsubsection{Feature extraction}
Videos are divided into non-overlapping 32-frame segments. We use the video swin transformer~\cite{liu2022video} for RGB frames and segmentation masks, and I3D~\cite{carreira2017quo} for optical flow, both pretrained on Kinetics-400, to extract visual features.
For training, we sample fixed-length consecutive segments from each dataset, with zero-padding applied when segments are shorter than the required length. All available segments are used during testing. Features from all modalities are projected to a dimension $d=256$ through separate embedding layers.

\subsubsection{Loss Function}
Following previous work~\cite{anastasiou2023keep}, we employ a hybrid loss function that combines Mean Squared Error (MSE) and Mean Absolute Error (MAE), which has demonstrated effectiveness in similar quality assessment tasks~\cite{parmar2019what}. Our loss function is defined as:
\begin{equation}
\mathcal{L} = \alpha \cdot \text{MSE}(\hat{y}, \bar{y}) + (1 - \alpha) \cdot \text{MAE}(\hat{y}, \bar{y})
\end{equation}

where $\hat{y}$ denotes the predicted score, $\bar{y}$ is the normalized label, and $\alpha$ is a hyperparameter that balances the two loss components. We set $\alpha$ to 0.5 throughout all experiments.

\begin{table*}[!ht]
\caption{Comparative results on the JIGSAWS dataset. K for kinematic, V for video, F for optical flow, M for segmentation mask, and SD for standard deviation.}
\label{tab:compare_table}
\centering
\renewcommand{\arraystretch}{1.2}
\setlength{\tabcolsep}{4.65pt}
\begin{tabular}{cccccccccccccc}
\hline
\multicolumn{14}{c}{\textbf{JIGSAWS}} \\ \hline
\multicolumn{1}{c|}{\multirow{2}{*}{\textbf{\begin{tabular}[c]{@{}c@{}}Input\\ Modality\end{tabular}}}} & \multicolumn{1}{c|}{\multirow{2}{*}{\textbf{Method}}} & \multicolumn{3}{c|}{\textbf{KT}} & \multicolumn{3}{c|}{\textbf{NP}} & \multicolumn{3}{c|}{\textbf{SU}} & \multicolumn{3}{c}{\textbf{Across Tasks}} \\ \cline{3-14} 
\multicolumn{1}{c|}{} & \multicolumn{1}{c|}{} & \multicolumn{1}{c|}{\textbf{LOSO}} & \multicolumn{1}{c|}{\textbf{LOUO}} & \multicolumn{1}{c|}{\textbf{4-Fold}} & \multicolumn{1}{c|}{\textbf{LOSO}} & \multicolumn{1}{c|}{\textbf{LOUO}} & \multicolumn{1}{c|}{\textbf{4-Fold}} & \multicolumn{1}{c|}{\textbf{LOSO}} & \multicolumn{1}{c|}{\textbf{LOUO}} & \multicolumn{1}{c|}{\textbf{4-Fold}} & \multicolumn{1}{c|}{\textbf{LOSO}} & \multicolumn{1}{c|}{\textbf{LOUO}} & \textbf{4-Fold} \\ \hline
\multicolumn{14}{c}{\textbf{Spearman's Correlation Coefficient (SSC)}} \\ \hline
\multicolumn{1}{c|}{\multirow{3}{*}{K}} & \multicolumn{1}{c|}{SMT-DCT-DFT\cite{zia2018automated}} & \multicolumn{1}{c|}{0.70} & \multicolumn{1}{c|}{0.73} & \multicolumn{1}{c|}{---} & \multicolumn{1}{c|}{0.38} & \multicolumn{1}{c|}{0.23} & \multicolumn{1}{c|}{---} & \multicolumn{1}{c|}{0.64} & \multicolumn{1}{c|}{0.10} & \multicolumn{1}{c|}{---} & \multicolumn{1}{c|}{0.59} & \multicolumn{1}{c|}{0.40} & --- \\
\multicolumn{1}{c|}{} & \multicolumn{1}{c|}{DCT-DFT-ApEn\cite{zia2018automated}} & \multicolumn{1}{c|}{0.63} & \multicolumn{1}{c|}{0.60} & \multicolumn{1}{c|}{---} & \multicolumn{1}{c|}{0.46} & \multicolumn{1}{c|}{0.25} & \multicolumn{1}{c|}{---} & \multicolumn{1}{c|}{0.75} & \multicolumn{1}{c|}{0.37} & \multicolumn{1}{c|}{---} & \multicolumn{1}{c|}{0.63} & \multicolumn{1}{c|}{0.42} & --- \\
\multicolumn{1}{c|}{} & \multicolumn{1}{c|}{ReCAP\cite{quarez2025recap}} & \multicolumn{1}{c|}{0.88} & \multicolumn{1}{c|}{---} & \multicolumn{1}{c|}{---} & \multicolumn{1}{c|}{0.85} & \multicolumn{1}{c|}{---} & \multicolumn{1}{c|}{---} & \multicolumn{1}{c|}{0.83} & \multicolumn{1}{c|}{---} & \multicolumn{1}{c|}{---} & \multicolumn{1}{c|}{0.85} & \multicolumn{1}{c|}{---} & --- \\ \hline
\multicolumn{1}{c|}{\multirow{6}{*}{V}} & \multicolumn{1}{c|}{C3D-LSTM\cite{parmar2017learning}} & \multicolumn{1}{c|}{0.81} & \multicolumn{1}{c|}{0.60} & \multicolumn{1}{c|}{---} & \multicolumn{1}{c|}{0.84} & \multicolumn{1}{c|}{0.78} & \multicolumn{1}{c|}{---} & \multicolumn{1}{c|}{0.69} & \multicolumn{1}{c|}{0.59} & \multicolumn{1}{c|}{---} & \multicolumn{1}{c|}{0.79} & \multicolumn{1}{c|}{0.67} & --- \\
\multicolumn{1}{c|}{} & \multicolumn{1}{c|}{C3D-MTL-VF\cite{wang2020towards}} & \multicolumn{1}{c|}{0.89} & \multicolumn{1}{c|}{0.83} & \multicolumn{1}{c|}{---} & \multicolumn{1}{c|}{0.75} & \multicolumn{1}{c|}{0.86} & \multicolumn{1}{c|}{---} & \multicolumn{1}{c|}{0.77} & \multicolumn{1}{c|}{0.69} & \multicolumn{1}{c|}{---} & \multicolumn{1}{c|}{0.81} & \multicolumn{1}{c|}{0.80} & --- \\
\multicolumn{1}{c|}{} & \multicolumn{1}{c|}{CoRe + GART\cite{yu2021group}} & \multicolumn{1}{c|}{---} & \multicolumn{1}{c|}{---} & \multicolumn{1}{c|}{0.86} & \multicolumn{1}{c|}{---} & \multicolumn{1}{c|}{---} & \multicolumn{1}{c|}{0.86} & \multicolumn{1}{c|}{---} & \multicolumn{1}{c|}{---} & \multicolumn{1}{c|}{0.84} & \multicolumn{1}{c|}{---} & \multicolumn{1}{c|}{---} & 0.85 \\
\multicolumn{1}{c|}{} & \multicolumn{1}{c|}{ViSA\cite{li2022surgical}} & \multicolumn{1}{c|}{{\underline{0.92}}} & \multicolumn{1}{c|}{0.76} & \multicolumn{1}{c|}{0.84} & \multicolumn{1}{c|}{\textbf{0.93}} & \multicolumn{1}{c|}{{\underline{0.90}}} & \multicolumn{1}{c|}{0.86} & \multicolumn{1}{c|}{0.84} & \multicolumn{1}{c|}{0.72} & \multicolumn{1}{c|}{0.79} & \multicolumn{1}{c|}{{\underline{0.90}}} & \multicolumn{1}{c|}{0.81} & 0.83 \\
\multicolumn{1}{c|}{} & \multicolumn{1}{c|}{Contra-Sformer\cite{anastasiou2023keep}} & \multicolumn{1}{c|}{0.89} & \multicolumn{1}{c|}{0.69} & \multicolumn{1}{c|}{0.87} & \multicolumn{1}{c|}{0.71} & \multicolumn{1}{c|}{0.71} & \multicolumn{1}{c|}{0.81} & \multicolumn{1}{c|}{0.86} & \multicolumn{1}{c|}{0.65} & \multicolumn{1}{c|}{0.69} & \multicolumn{1}{c|}{0.83} & \multicolumn{1}{c|}{0.68} & 0.81 \\
\multicolumn{1}{c|}{} & \multicolumn{1}{c|}{CoRe + PECoP\cite{dadashzadeh2024pecop}} & \multicolumn{1}{c|}{---} & \multicolumn{1}{c|}{---} & \multicolumn{1}{c|}{{\underline{0.88}}} & \multicolumn{1}{c|}{---} & \multicolumn{1}{c|}{---} & \multicolumn{1}{c|}{\textbf{0.90}} & \multicolumn{1}{c|}{---} & \multicolumn{1}{c|}{---} & \multicolumn{1}{c|}{{\underline{0.88}}} & \multicolumn{1}{c|}{---} & \multicolumn{1}{c|}{---} & {\underline{0.89}} \\ \hline
\multicolumn{1}{c|}{\multirow{3}{*}{V+K}} & \multicolumn{1}{c|}{AIM\cite{gao2020asymmetric}} & \multicolumn{1}{c|}{---} & \multicolumn{1}{c|}{0.61} & \multicolumn{1}{c|}{0.82} & \multicolumn{1}{c|}{---} & \multicolumn{1}{c|}{0.34} & \multicolumn{1}{c|}{0.65} & \multicolumn{1}{c|}{---} & \multicolumn{1}{c|}{0.45} & \multicolumn{1}{c|}{0.63} & \multicolumn{1}{c|}{---} & \multicolumn{1}{c|}{0.47} & 0.71 \\
\multicolumn{1}{c|}{} & \multicolumn{1}{c|}{MultiPath-VTP\cite{liu2021towards}} & \multicolumn{1}{c|}{---} & \multicolumn{1}{c|}{0.58} & \multicolumn{1}{c|}{0.78} & \multicolumn{1}{c|}{---} & \multicolumn{1}{c|}{0.62} & \multicolumn{1}{c|}{0.76} & \multicolumn{1}{c|}{---} & \multicolumn{1}{c|}{0.45} & \multicolumn{1}{c|}{0.79} & \multicolumn{1}{c|}{---} & \multicolumn{1}{c|}{0.55} & 0.78 \\
\multicolumn{1}{c|}{} & \multicolumn{1}{c|}{MultiPath-VTPE\cite{liu2021towards}} & \multicolumn{1}{c|}{---} & \multicolumn{1}{c|}{0.59} & \multicolumn{1}{c|}{0.82} & \multicolumn{1}{c|}{---} & \multicolumn{1}{c|}{0.65} & \multicolumn{1}{c|}{0.76} & \multicolumn{1}{c|}{---} & \multicolumn{1}{c|}{0.45} & \multicolumn{1}{c|}{0.83} & \multicolumn{1}{c|}{---} & \multicolumn{1}{c|}{0.57} & 0.81 \\ \hline
\multicolumn{1}{c|}{\multirow{3}{*}{V+F+M}} & \multicolumn{1}{c|}{MAQA\cite{zeng2024multimodal}} & \multicolumn{1}{c|}{0.90} & \multicolumn{1}{c|}{{\underline{0.86}}} & \multicolumn{1}{c|}{0.87} & \multicolumn{1}{c|}{0.89} & \multicolumn{1}{c|}{0.87} & \multicolumn{1}{c|}{0.87} & \multicolumn{1}{c|}{{\underline{0.87}}} & \multicolumn{1}{c|}{{\underline{0.82}}} & \multicolumn{1}{c|}{0.86} & \multicolumn{1}{c|}{0.89} & \multicolumn{1}{c|}{{\underline{0.85}}} & 0.87 \\
\multicolumn{1}{c|}{} & \multicolumn{1}{c|}{SurgFusion-Net (Ours)} & \multicolumn{1}{c|}{\textbf{0.93}} & \multicolumn{1}{c|}{\textbf{0.91}} & \multicolumn{1}{c|}{\textbf{0.91}} & \multicolumn{1}{c|}{{\underline{0.92}}} & \multicolumn{1}{c|}{\textbf{0.91}} & \multicolumn{1}{c|}{{\underline{0.89}}} & \multicolumn{1}{c|}{\textbf{0.91}} & \multicolumn{1}{c|}{\textbf{0.84}} & \multicolumn{1}{c|}{\textbf{0.89}} & \multicolumn{1}{c|}{\textbf{0.92}} & \multicolumn{1}{c|}{\textbf{0.89}} & \textbf{0.90} \\
\multicolumn{1}{c|}{} & \multicolumn{1}{c|}{SD of SCC} & \multicolumn{1}{c|}{±0.076} & \multicolumn{1}{c|}{±0.118} & \multicolumn{1}{c|}{±0.058} & \multicolumn{1}{c|}{±0.092} & \multicolumn{1}{c|}{±0.119} & \multicolumn{1}{c|}{±0.094} & \multicolumn{1}{c|}{±0.032} & \multicolumn{1}{c|}{±0.132} & \multicolumn{1}{c|}{±0.090} & \multicolumn{1}{c|}{±0.071} & \multicolumn{1}{c|}{±0.123} & ±0.082 \\ \hline
\multicolumn{14}{c}{\textbf{Mean Absolute Error (MAE)}} \\ \hline
\multicolumn{1}{c|}{\multirow{2}{*}{V}} & \multicolumn{1}{c|}{ViSA\cite{li2022surgical}} & \multicolumn{1}{c|}{2.16} & \multicolumn{1}{c|}{2.01} & \multicolumn{1}{c|}{2.60} & \multicolumn{1}{c|}{{\underline{1.66}}} & \multicolumn{1}{c|}{2.16} & \multicolumn{1}{c|}{\textbf{2.05}} & \multicolumn{1}{c|}{2.58} & \multicolumn{1}{c|}{2.82} & \multicolumn{1}{c|}{2.70} & \multicolumn{1}{c|}{2.13} & \multicolumn{1}{c|}{2.33} & {\underline{2.45}} \\
\multicolumn{1}{c|}{} & \multicolumn{1}{c|}{Contra-Sformer\cite{anastasiou2023keep}} & \multicolumn{1}{c|}{\textbf{1.75}} & \multicolumn{1}{c|}{{\underline{1.39}}} & \multicolumn{1}{c|}{\textbf{2.10}} & \multicolumn{1}{c|}{3.15} & \multicolumn{1}{c|}{3.17} & \multicolumn{1}{c|}{3.21} & \multicolumn{1}{c|}{2.74} & \multicolumn{1}{c|}{2.58} & \multicolumn{1}{c|}{2.99} & \multicolumn{1}{c|}{2.55} & \multicolumn{1}{c|}{2.38} & 2.77 \\ \hline
\multicolumn{1}{c|}{\multirow{3}{*}{V+F+M}} & \multicolumn{1}{c|}{MAQA\cite{zeng2024multimodal}} & \multicolumn{1}{c|}{2.06} & \multicolumn{1}{c|}{1.68} & \multicolumn{1}{c|}{2.47} & \multicolumn{1}{c|}{1.82} & \multicolumn{1}{c|}{{\underline{2.05}}} & \multicolumn{1}{c|}{2.53} & \multicolumn{1}{c|}{{\underline{2.28}}} & \multicolumn{1}{c|}{{\underline{1.69}}} & \multicolumn{1}{c|}{{\underline{2.60}}} & \multicolumn{1}{c|}{{\underline{2.05}}} & \multicolumn{1}{c|}{{\underline{1.81}}} & 2.53 \\
\multicolumn{1}{c|}{} & \multicolumn{1}{c|}{SurgFusion-Net (Ours)} & \multicolumn{1}{c|}{{\underline{1.88}}} & \multicolumn{1}{c|}{\textbf{1.19}} & \multicolumn{1}{c|}{{\underline{2.30}}} & \multicolumn{1}{c|}{\textbf{1.49}} & \multicolumn{1}{c|}{\textbf{1.66}} & \multicolumn{1}{c|}{{\underline{2.37}}} & \multicolumn{1}{c|}{\textbf{2.13}} & \multicolumn{1}{c|}{\textbf{1.49}} & \multicolumn{1}{c|}{\textbf{2.53}} & \multicolumn{1}{c|}{\textbf{1.83}} & \multicolumn{1}{c|}{\textbf{1.45}} & \textbf{2.40} \\
\multicolumn{1}{c|}{} & \multicolumn{1}{c|}{SD of MAE} & \multicolumn{1}{c|}{±0.481} & \multicolumn{1}{c|}{±0.357} & \multicolumn{1}{c|}{±0.675} & \multicolumn{1}{c|}{±0.519} & \multicolumn{1}{c|}{±0.813} & \multicolumn{1}{c|}{±0.565} & \multicolumn{1}{c|}{±0.663} & \multicolumn{1}{c|}{±0.776} & \multicolumn{1}{c|}{±0.588} & \multicolumn{1}{c|}{±0.560} & \multicolumn{1}{c|}{±0.681} & ±0.611 \\ \hline
\end{tabular}
\end{table*}

\section{EXPERIMENTS}
\subsection{Datasets and Evaluation Metrics}
In addition to our RAH-skill and RARP-skill datasets, we evaluate SurgFusion-Net on the JIGSAWS benchmark~\cite{ahmidi2017dataset} using all three tasks, suturing (SU), needle passing (NP), and knot tying (KT). The dataset contains 39 SU, 28 NP, and 36 KT videos from eight surgeons (2 experts, 2 intermediate, 4 novices), with GRS scores ranging from 6-30.
Videos are sampled at 10 FPS with optical flow extracted using Farnebäck algorithm and segmentation masks generated via SAM 2~\cite{ravi2025sam}.
For segmentation, we annotated surgical instruments and task-specific objects: sutures and reference elements for SU, yellow tube and sutures for KT, and reference objects for NP, as shown in Fig.~\ref{fig:datasets}(c)-(e). This yields 43,914 masks for SU, 20,623 masks for KT, and 30,376 masks for NP.

Following previous work~\cite{zeng2024multimodal},~\cite{li2022surgical}, we use Spearman's Correlation Coefficient (SCC) and Mean Absolute Error (MAE) as evaluation metrics. 
SCC ranges from -1 to 1 and measures ranking correlation, while MAE measures prediction accuracy, with higher SCC and lower MAE indicating better performance.
We employ cross-validation for both unimodal and multimodal training. In JIGSAWS, we use the predefined Leave-One-Supertrial-Out (LOSO) and Leave-One-User-Out (LOUO)~\cite{ahmidi2017dataset}, and random 4-Fold. For RAH-skill and RARP-skill, we use random 4-Fold cross-validation. For each scheme, we calculate the mean SCC across all folds and obtain the SCC across JIGSAWS tasks through Fisher's z-transformation~\cite{liu2021towards}. We report the best SCC and its corresponding MAE for each cross-validation scheme.

\begin{table}[!ht]
\caption{Comparative results on RAH-skill and RARP-skill datasets.}
\label{tab:compare_table_vvs}
\centering
\renewcommand{\arraystretch}{1.2}
\setlength{\tabcolsep}{3.76pt}
\begin{tabular}{c|c|cc|cc}
\hline
\multirow{2}{*}{\textbf{\begin{tabular}[c]{@{}c@{}}Input\\ Modality\end{tabular}}} & \multirow{2}{*}{\textbf{Method}} & \multicolumn{2}{c|}{\textbf{RAH-skill}} & \multicolumn{2}{c}{\textbf{RARP-skill}} \\ \cline{3-6} 
 &  & \multicolumn{1}{c|}{\textbf{SCC}} & \multicolumn{1}{c|}{\textbf{MAE}} & \multicolumn{1}{c|}{\textbf{SCC}} & \textbf{MAE} \\ \hline
\multirow{4}{*}{V} & CoRe + GART\cite{yu2021group} & \multicolumn{1}{c|}{0.5893} & \multicolumn{1}{c|}{2.9434} & \multicolumn{1}{c|}{0.5656} & 2.0240 \\
 & ViSA\cite{li2022surgical} & \multicolumn{1}{c|}{0.5856} & \multicolumn{1}{c|}{2.6852} & \multicolumn{1}{c|}{0.5997} & 2.1041 \\
 & Contra-Sformer\cite{anastasiou2023keep} & \multicolumn{1}{c|}{0.5695} & \multicolumn{1}{c|}{2.5428} & \multicolumn{1}{c|}{0.6083} & 1.6345 \\
 & CoRe + PECoP\cite{dadashzadeh2024pecop} & \multicolumn{1}{c|}{0.6658} & \multicolumn{1}{c|}{2.8975} & \multicolumn{1}{c|}{0.7103} & 2.2799 \\ \hline
\multirow{3}{*}{V+F+M} & MAQA\cite{zeng2024multimodal} & \multicolumn{1}{c|}{{\underline{0.7756}}} & \multicolumn{1}{c|}{{\underline{1.9645}}} & \multicolumn{1}{c|}{{\underline{0.7993}}} & {\underline{1.3588}} \\
 & SurgFus-Net (Ours) & \multicolumn{1}{c|}{\textbf{0.8294}} & \multicolumn{1}{c|}{\textbf{1.8182}} & \multicolumn{1}{c|}{\textbf{0.8486}} & \textbf{1.2974} \\
 & Standard Deviation & \multicolumn{1}{c|}{±0.0250} & \multicolumn{1}{c|}{±0.3066} & \multicolumn{1}{c|}{±0.0245} & ±0.4222 \\ \hline
\end{tabular}
\end{table}

\subsection{Implementation Details}
Since the multimodal fusion branch is based on pretrained unimodal branches, SurgFusion-Net is trained in two phases following the training protocol in~\cite{zeng2024multimodal}.
In the first phase, unimodal branches are optimized using SGD with momentum 0.9 and weight decay $10^{-4}$. The learning rate follows a cosine decay schedule from $10^{-2}$ to $5 \times 10^{-6}$, with batch size 32 for RAH-skill and 16 for RARP-skill and JIGSAWS. Training runs for 300 epochs to obtain pretrained weights.
In the second phase, unimodal branches are frozen and their regression heads are disabled. The multimodal fusion branch is trained with AdamW~\cite{loshchilov2019decoupled} using a cosine decay learning rate schedule from $10^{-3}$ to $5 \times 10^{-6}$. Training uses the same batch sizes and runs for 300 epochs. The number of FusionNets K is set to 10 following~\cite{zeng2024multimodal}. The model is implemented in PyTorch and trained on a RTX 4090 GPU. 

\begin{table*}[!ht]
\caption{Evaluation of the fusion branches on JIGSAWS, RAH-skill and RARP-skill datasets.}
\label{tab:ablation_branch}
\centering
\renewcommand{\arraystretch}{1.2}
\setlength{\tabcolsep}{5.5pt}
\begin{tabular}{c|ccc|cccccc|cc|cc}
\hline
\multirow{3}{*}{\textbf{\begin{tabular}[c]{@{}c@{}}Branch\\ (Block)\end{tabular}}} & \multicolumn{3}{c|}{\multirow{2}{*}{\textbf{Modality}}} & \multicolumn{6}{c|}{\textbf{JIGSAWS (LOUO)}} & \multicolumn{2}{c|}{\multirow{2}{*}{\textbf{RAH-skill}}} & \multicolumn{2}{c}{\multirow{2}{*}{\textbf{RARP-skill}}} \\ \cline{5-10}
 & \multicolumn{3}{c|}{} & \multicolumn{2}{c|}{\textbf{KT}} & \multicolumn{2}{c|}{\textbf{NP}} & \multicolumn{2}{c|}{\textbf{SU}} & \multicolumn{2}{c|}{} & \multicolumn{2}{c}{} \\ \cline{2-14} 
 & \multicolumn{1}{c|}{\textbf{RGB}} & \multicolumn{1}{c|}{\textbf{Flow}} & \textbf{Mask} & \multicolumn{1}{c|}{\textbf{SCC}} & \multicolumn{1}{c|}{\textbf{MAE}} & \multicolumn{1}{c|}{\textbf{SCC}} & \multicolumn{1}{c|}{\textbf{MAE}} & \multicolumn{1}{c|}{\textbf{SCC}} & \textbf{MAE} & \multicolumn{1}{c|}{\textbf{SCC}} & \textbf{MAE} & \multicolumn{1}{c|}{\textbf{SCC}} & \textbf{MAE} \\ \hline
\multirow{3}{*}{\begin{tabular}[c]{@{}c@{}}Single \\ Modality\end{tabular}} & \multicolumn{1}{c|}{\checkmark} & \multicolumn{1}{c|}{} &  & \multicolumn{1}{c|}{0.8422} & \multicolumn{1}{c|}{1.7873} & \multicolumn{1}{c|}{0.8624} & \multicolumn{1}{c|}{2.5825} & \multicolumn{1}{c|}{0.7748} & 1.9907 & \multicolumn{1}{c|}{0.6960} & 2.3041 & \multicolumn{1}{c|}{0.7368} & 2.0024 \\
 & \multicolumn{1}{c|}{} & \multicolumn{1}{c|}{\checkmark} &  & \multicolumn{1}{c|}{0.8307} & \multicolumn{1}{c|}{2.0367} & \multicolumn{1}{c|}{0.8431} & \multicolumn{1}{c|}{2.8785} & \multicolumn{1}{c|}{0.7539} & 1.8390 & \multicolumn{1}{c|}{0.6282} & 2.8250 & \multicolumn{1}{c|}{0.7254} & 1.8059 \\
 & \multicolumn{1}{c|}{} & \multicolumn{1}{c|}{} & \checkmark & \multicolumn{1}{c|}{0.8367} & \multicolumn{1}{c|}{1.5216} & \multicolumn{1}{c|}{0.8315} & \multicolumn{1}{c|}{2.1611} & \multicolumn{1}{c|}{0.7663} & 1.7851 & \multicolumn{1}{c|}{0.7056} & 2.1335 & \multicolumn{1}{c|}{0.7459} & 1.7138 \\ \hline
\multirow{4}{*}{\begin{tabular}[c]{@{}c@{}}Multimodal\\ Fusion\\ (CSFB)\end{tabular}} & \multicolumn{1}{c|}{\checkmark} & \multicolumn{1}{c|}{\checkmark} &  & \multicolumn{1}{c|}{0.8783} & \multicolumn{1}{c|}{1.4713} & \multicolumn{1}{c|}{0.8746} & \multicolumn{1}{c|}{2.0201} & \multicolumn{1}{c|}{0.8044} & 1.7529 & \multicolumn{1}{c|}{0.7441} & 2.0762 & \multicolumn{1}{c|}{0.7891} & 1.4823 \\
 & \multicolumn{1}{c|}{\checkmark} & \multicolumn{1}{c|}{} & \checkmark & \multicolumn{1}{c|}{0.8813} & \multicolumn{1}{c|}{1.4215} & \multicolumn{1}{c|}{0.8721} & \multicolumn{1}{c|}{1.8586} & \multicolumn{1}{c|}{0.8103} & 1.6957 & \multicolumn{1}{c|}{0.7498} & 2.0342 & \multicolumn{1}{c|}{0.7978} & 1.4614 \\
 & \multicolumn{1}{c|}{} & \multicolumn{1}{c|}{\checkmark} & \checkmark & \multicolumn{1}{c|}{0.8685} & \multicolumn{1}{c|}{1.4387} & \multicolumn{1}{c|}{0.8657} & \multicolumn{1}{c|}{1.8830} & \multicolumn{1}{c|}{0.7983} & 1.7266 & \multicolumn{1}{c|}{0.7269} & 2.0564 & \multicolumn{1}{c|}{0.7712} & 1.5048 \\
 & \multicolumn{1}{c|}{\checkmark} & \multicolumn{1}{c|}{\checkmark} & \checkmark & \multicolumn{1}{c|}{{0.8976}} & \multicolumn{1}{c|}{1.4344} & \multicolumn{1}{c|}{{0.8876}} & \multicolumn{1}{c|}{1.8383} & \multicolumn{1}{c|}{0.8178} & 1.6722 & \multicolumn{1}{c|}{0.7557} & 2.0356 & \multicolumn{1}{c|}{0.8082} & 1.4883 \\ \hline
\multirow{4}{*}{\begin{tabular}[c]{@{}c@{}}Multimodal\\ Fusion\\ (CSFB + DFB)\end{tabular}} & \multicolumn{1}{c|}{\checkmark} & \multicolumn{1}{c|}{\checkmark} &  & \multicolumn{1}{c|}{0.9051} & \multicolumn{1}{c|}{1.3326} & \multicolumn{1}{c|}{\underline{0.9012}} & \multicolumn{1}{c|}{1.7206} & \multicolumn{1}{c|}{\underline{0.8259}} & 1.5574 & \multicolumn{1}{c|}{0.7643} & {2.0125} & \multicolumn{1}{c|}{0.8109} & 1.4175 \\
 & \multicolumn{1}{c|}{\checkmark} & \multicolumn{1}{c|}{} & \checkmark & \multicolumn{1}{c|}{\underline{0.9064}} & \multicolumn{1}{c|}{1.2856} & \multicolumn{1}{c|}{0.8994} & \multicolumn{1}{c|}{{\underline{1.6674}}} & \multicolumn{1}{c|}{0.8227} & \underline{1.4957} & \multicolumn{1}{c|}{{\underline{0.7732}}} & \underline{1.9272} & \multicolumn{1}{c|}{{\underline{0.8210}}} & {\underline{1.3681}} \\
 & \multicolumn{1}{c|}{} & \multicolumn{1}{c|}{\checkmark} & \checkmark & \multicolumn{1}{c|}{0.8826} & \multicolumn{1}{c|}{{\underline{1.2749}}} & \multicolumn{1}{c|}{0.8784} & \multicolumn{1}{c|}{1.6852} & \multicolumn{1}{c|}{0.8023} & {1.5280} & \multicolumn{1}{c|}{0.7434} & 2.1004 & \multicolumn{1}{c|}{0.8037} & 1.3821 \\
 & \multicolumn{1}{c|}{\checkmark} & \multicolumn{1}{c|}{\checkmark} & \checkmark & \multicolumn{1}{c|}{\textbf{0.9135}} & \multicolumn{1}{c|}{\textbf{1.1929}} & \multicolumn{1}{c|}{\textbf{0.9092}} & \multicolumn{1}{c|}{\textbf{1.6581}} & \multicolumn{1}{c|}{\textbf{0.8355}} & \textbf{1.4866} & \multicolumn{1}{c|}{\textbf{0.8294}} & \textbf{1.8182} & \multicolumn{1}{c|}{\textbf{0.8486}} & \textbf{1.2974} \\ \hline
\end{tabular}
\end{table*}

\subsection{Experimental Results}
We compare SurgFusion-Net with existing SOTA RAS skill assessment and AQA methods. We reimplement MAQA ~\cite{zeng2024multimodal} by replacing the audio modality with segmentation masks to enable fair comparison.
As shown in Table~\ref{tab:compare_table}, our method achieves the highest average SCC scores across tasks on JIGSAWS, improving by 0.02 on LOSO and 0.04 on LOUO over the second best methods.
Compared with the previous best-performing RAS skill assessment method~\cite{li2022surgical}, SurgFusion-Net achieves significant improvements of 0.02 and 0.08 in average SCC on JIGSAWS for LOSO and LOUO schemes, while reducing MAE by 0.30 and 0.88, indicating that multimodal fusion outperforms RGB-only approaches. 
Compared with multimodal AQA methods~\cite{zeng2024multimodal, liu2021towards}, our approach better exploits multimodal information, achieving improvements of 0.03 and 0.09 in across-task average SCC over MAQA and MultiPath-VTPE on JIGSAWS under random 4-fold evaluation, with 0.13 MAE reduction compared to MAQA.
This demonstrates the effectiveness of SurgFusion-Net, where the combination of RGB, optical flow, and segmentation masks outperforms RGB-kinematics fusion due to better dimensionality consistency that facilitates cross-modal fusion.

As shown in Table~\ref{tab:compare_table_vvs}, SurgFusion-Net achieves the best average SCC and MAE under random 4-fold cross validation on RAH-skill and RARP-skill. The SOTA video-based AQA and surgical skill assessment methods~\cite{yu2021group, li2022surgical, anastasiou2023keep} analyze surgical actions by extracting key frames, achieving approximately 0.6 average SCC on both datasets. This limitation in performance occurs because clinical videos are longer than the JIGSAWS training videos, exhibit frequent camera movements, and involve domain shift, making it challenging to transfer general AQA methods to surgical skill assessment.
By conducting self-supervised pre-training on RAH-skill and RARP-skill, CoRe+PECoP~\cite{dadashzadeh2024pecop} improves SCC by 0.077 and 0.102 over CoRe+GART~\cite{yu2021group} and Contra-Sformer~\cite{anastasiou2023keep}, demonstrating that pre-training improves score estimation on clinical datasets. Both MAQA and SurgFusion-Net implement stepwise training, with our model outperforming MAQA by 0.054 and 0.049 in average SCC. This demonstrates the superiority of DRA for multimodal fusion compared to the recent state-of-the-art.
Moreover, our model reduces average MAE by 0.725 on RAH-skill and 0.337 on RARP-skill compared to the RGB-only method~\cite{anastasiou2023keep}, highlighting the improved accuracy achieved through multimodal fusion.

\begin{table}[!ht]
\caption{Evaluation of the fusion Strategy on JIGSAWS (SU), RAH-skill and RARP-skill datasets. CA for Cross-Attention. }
\label{tab:ablation_fusion_method}
\centering
\renewcommand{\arraystretch}{1.2}
\setlength{\tabcolsep}{4pt}
\begin{tabular}{c|cc|cc|cc}
\hline
\multirow{2}{*}{\textbf{\begin{tabular}[c]{@{}c@{}}Fusion \\ Strategy\end{tabular}}} & \multicolumn{2}{c|}{\textbf{SU (LOUO)}} & \multicolumn{2}{c|}{\textbf{RAH-skill}} & \multicolumn{2}{c}{\textbf{RARP-skill}} \\ \cline{2-7} 
 & \multicolumn{1}{c|}{\textbf{SCC}} & \textbf{MAE} & \multicolumn{1}{c|}{\textbf{SCC}} & \textbf{MAE} & \multicolumn{1}{c|}{\textbf{SCC}} & \textbf{MAE} \\ \hline
AVG & \multicolumn{1}{c|}{0.8081} & {\underline{1.5491}} & \multicolumn{1}{c|}{0.7529} & 1.9340 & \multicolumn{1}{c|}{0.7379} & 1.3910 \\
CAT & \multicolumn{1}{c|}{0.8110} & 1.9093 & \multicolumn{1}{c|}{0.7633} & 1.9981 & \multicolumn{1}{c|}{0.7460} & 1.4304 \\
Vanilla CA\cite{vaswani2017attention} & \multicolumn{1}{c|}{0.8154} & 1.8795 & \multicolumn{1}{c|}{0.7676} & {\underline{1.9031}} & \multicolumn{1}{c|}{0.7853} & {\underline{1.3102}} \\
Negative CA\cite{zeng2024multimodal} & \multicolumn{1}{c|}{{\underline{0.8210}}} & 1.6883 & \multicolumn{1}{c|}{{\underline{0.7756}}} & 1.9645 & \multicolumn{1}{c|}{{\underline{0.7993}}} & 1.3588 \\
DRA (Ours) & \multicolumn{1}{c|}{\textbf{0.8355}} & \textbf{1.4866} & \multicolumn{1}{c|}{\textbf{0.8294}} & \textbf{1.8182} & \multicolumn{1}{c|}{\textbf{0.8486}} & \textbf{1.2974} \\ \hline
\end{tabular}
\end{table}

\subsection{Ablation Studies}
\subsubsection{Evaluation of the fusion modules}
To validate the contribution of each module, we test unimodal and multimodal fusion branches on JIGSAWS (LOUO scheme), RAH-skill, and RARP-skill. For isolated testing of the CSFB block, we mask the DFB block using zero-initialized placeholders.
As shown in Table~\ref{tab:ablation_branch}, although the unimodal branch uses only three residual convolutional blocks, it achieves 0.86 on JIGSAWS NP task with RGB input, matching C3D-MTL-VF~\cite{wang2020towards} performance. The unimodal branch using optical flow shows poor performance on RAH-skill with average SCC of 0.6282, likely due to degraded optical flow quality from frequent camera adjustments and tissue motion during vaginal vault closure.
The multimodal fusion branch outperforms unimodal branches, achieving higher SCC across all datasets. Specifically, using CSFB to fuse three-modal information yields SCC improvements of 0.055 and 0.062 over the RGB-based unimodal branch on JIGSAWS KT and RARP-skill. Adding DFB further enhances performance by 0.016 and 0.040, respectively. These results justify the architecture of SurgFusion-Net, and the greater improvements observed in the two clinical datasets highlight its potential for clinical adoption.
MAE exhibits an decreasing trend, with the reduction being slower on RAH-skill, likely due to increased noise in RGB and optical flow data, which impedes MAE optimization.

\subsubsection{Evaluation of Fusion Strategies}
We demonstrate the superiority of our DRA fusion strategy, and compare it with: averaging (AVG), concatenation (CAT), vanilla cross-attention (CA)~\cite{vaswani2017attention}, and negative cross-attention~\cite{zeng2024multimodal}.
For AVG, we calculate the average of the three modality features, while for CAT, we concatenate the features and use an additional fully connected (FC) layer for dimension reduction. We implement vanilla CA following~\cite{vaswani2017attention} and negative CA following~\cite{zeng2024multimodal}, directly replacing DRA with these methods.
\begin{table}[!ht]
\caption{Evaluation of the number of heads on JIGSAWS (SU), RAH-skill and RARP-skill datasets.}
\label{tab:ablation_num_head}
\centering
\renewcommand{\arraystretch}{1.2}
\setlength{\tabcolsep}{5.5pt}
\begin{tabular}{c|cc|cc|cc}
\hline
\multirow{2}{*}{\textbf{\# Heads}} & \multicolumn{2}{c|}{\textbf{SU (LOUO)}} & \multicolumn{2}{c|}{\textbf{RAH-skill}} & \multicolumn{2}{c}{\textbf{RARP-skill}} \\ \cline{2-7} 
 & \multicolumn{1}{c|}{\textbf{SCC}} & \textbf{MAE} & \multicolumn{1}{c|}{\textbf{SCC}} & \textbf{MAE} & \multicolumn{1}{c|}{\textbf{SCC}} & \textbf{MAE} \\ \hline
1 & \multicolumn{1}{c|}{0.8223} & 1.5195 & \multicolumn{1}{c|}{0.7859} & {\underline{1.8433}} & \multicolumn{1}{c|}{0.8055} & {\underline{1.3298}} \\
2 & \multicolumn{1}{c|}{\textbf{0.8355}} & \textbf{1.4866} & \multicolumn{1}{c|}{\textbf{0.8294}} & \textbf{1.8182} & \multicolumn{1}{c|}{\textbf{0.8486}} & \textbf{1.2974} \\
4 & \multicolumn{1}{c|}{{\underline{0.8279}}} & {\underline{1.4954}} & \multicolumn{1}{c|}{0.7892} & 1.8875 & \multicolumn{1}{c|}{0.8246} & 1.3375 \\
8 & \multicolumn{1}{c|}{0.8221} & 1.7481 & \multicolumn{1}{c|}{{\underline{0.7927}}} & 1.9112 & \multicolumn{1}{c|}{{\underline{0.8270}}} & 1.3717 \\ \hline
\end{tabular}
\end{table}
Table~\ref{tab:ablation_fusion_method} shows that DRA fusion outperforms all methods across three datasets, with SCC improvements of 0.054 and 0.049 on RAH-skill and RARP-skill respectively. DRA also achieves the lowest MAE, with reductions of 0.063 and 0.085 compared to AVG on JIGSAWS (SU) and vanilla CA on RAH-skill. The consistent improvements across simulation and clinical datasets indicate that the DRA mechanism effectively handles diverse surgical video content. 
These results highlight the benefits of DRA for multimodal fusion,  particularly in the challenging video regression task of surgical skill assessment.

\begin{figure*}[!ht]
    \centering
    \includegraphics[width=1.0\textwidth]
    {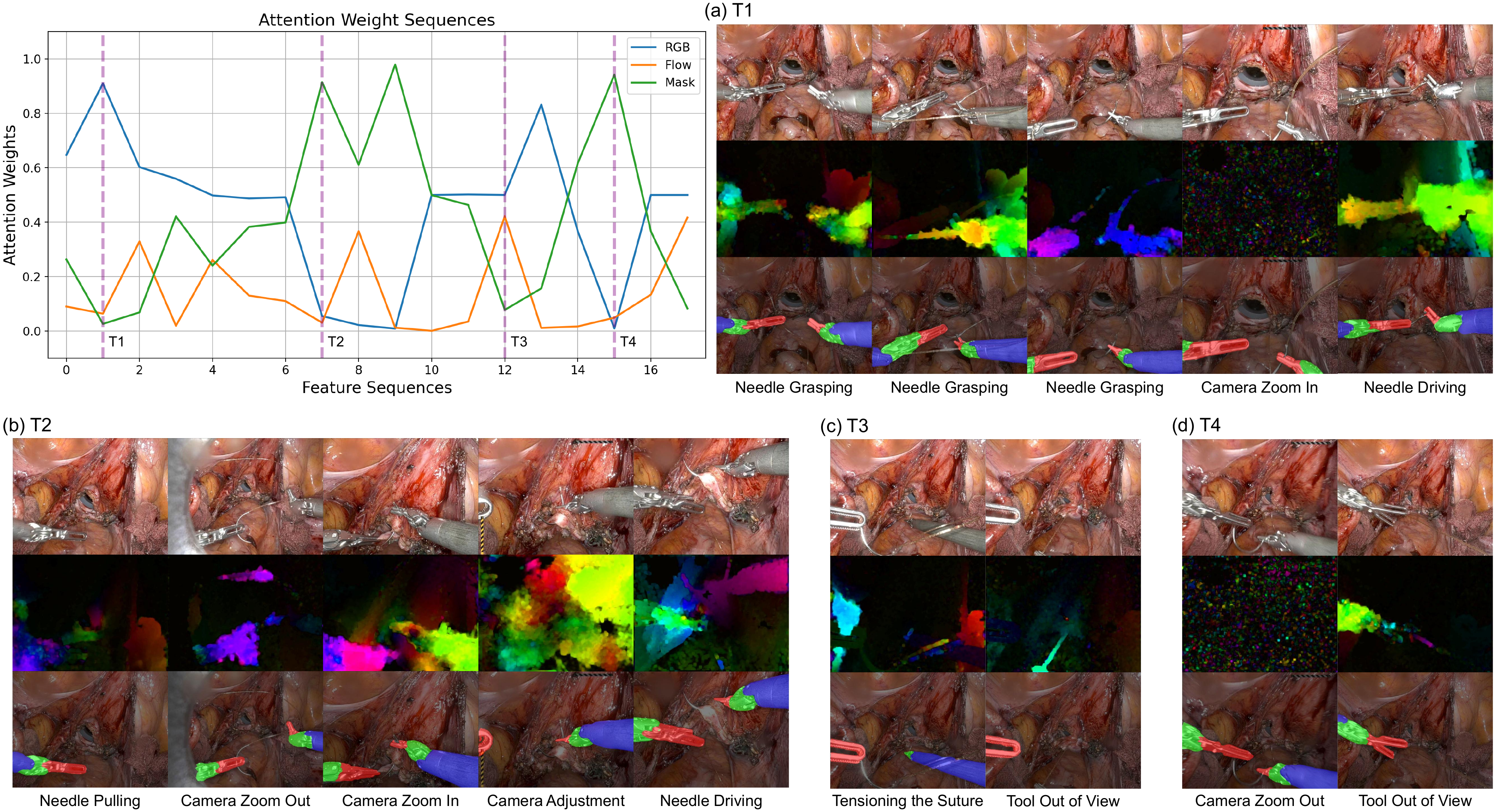}
    \caption{Visualization of attention weights of the cross-stage fusion block (CSFB) in the stage two on RAH-skill dataset. The top left figure displays the attention weights across three modalities on feature sequences, with four purple vertical dashed lines marking the time windows of specific features. Figures (a)-(d) show RGB frames, optical flows, and segmentation masks (from top to bottom in each column) corresponding to time windows T1 to T4. In stage two, each horizontal interval corresponds to about 14 seconds of video.}
    \label{fig:quali_fig}
\end{figure*}

\subsubsection{Evaluation of DRA Components}
We also investigate DRA's adaptive mixed attention (Eqs.\ref{eq:A_pls}--\ref{eq:A_mix}) and diversified multi-head attention (Eqs.\ref{eq:main_pattern}--\ref{eq:A_final}), by evaluating performance using different numbers of attention heads on JIGSAWS (SU), RAH-skill, and RARP-skill. The first column in Table~\ref{tab:ablation_num_head} indicates the number of attention heads in CSFB and DFB. Under single-head configuration, the diversification mechanism (Eqs.~\ref{eq:main_pattern}--\ref{eq:A_final}) becomes inactive, reducing to $\mathcal{A}_h^{final} = \mathcal{A}_h^{mix}$.
Compared to the baseline methods from Table~\ref{tab:ablation_fusion_method}, the single-head SurgFusion-Net (Table~\ref{tab:ablation_num_head}) shows improvements of 0.018 and 0.010 over vanilla CA and negative CA respectively on RAH-skill, demonstrating the advantages of adaptive mixed attention.
As shown in Table~\ref{tab:ablation_num_head}, SurgFusion-Net achieves optimal performance with dual attention heads across all three datasets for both average SCC and MAE. Compared to single-head, diversified dual-head attention improves SCC by 0.013 on SU. The improvement is further amplified on RAH-skill and RARP-skill, reaching 0.044 and 0.043 respectively. Across all three datasets, quad-head attention consistently outperforms single-head attention, with an improvement of 0.019 on RARP-skill, further validating the effectiveness of diversified multi-head attention. 
With eight attention heads, the limited capacity ($d_h=32$) leads to increased MAE.

\subsection{Qualitative Results}
To provide insights on the significance of the different modalities learned by our method, we visualize the attention weights of the CSFB on a video feature sequence from stage two, together with the corresponding video frames, optical flow, and segmentation masks, as shown in Fig.~\ref{fig:quali_fig}.
Prior to T1, during needle grasping, RGB exhibits the highest attention weights (0.64), exceeding segmentation masks (0.27) and optical flow (0.09). The camera zoom-in operation temporarily compromises the reliability of optical flow calculation, while surgical tools are positioned at the periphery, resulting in diminished attention weights for both modalities. During needle driving under magnified view, intensified background noise causes RGB and optical flow weights to decrease, while centrally located tools increase segmentation mask weights. 
During T2 (Fig.~\ref{fig:quali_fig}(b)), more camera adjustments occur. When surgical tools move out of view or become smaller during needle pulling and zoom-out, tool segmentation weights decrease while optical flow captures instrument and suture movements with increased attention. Camera zoom-in and view adjustments maintain RGB and optical flow at low levels, with segmentation masks dominating. During needle driving under centered view, both RGB and segmentation masks achieve predominant weights (around 0.5).
At T3 (Fig.~\ref{fig:quali_fig}(c)), when tensioning the suture, surgical tools move out of the visual field, causing segmentation weights to drop to 0.08, while RGB (0.51) and optical flow (0.41) become dominant.
Overall, segmentation mask weights decrease when tools are at field edges or out of view. Camera adjustments temporarily disrupt optical flow, which also suffers from background noise under magnified view. Rapid camera movements create motion artifacts that reduce RGB weights. The attention weight curves in Fig.~\ref{fig:quali_fig} demonstrate that our SurgFusion-Net with DRA can adaptively fuse multimodal features based on surgical actions and scene variations, enhancing surgical skill assessment accuracy. More visualization results are provided in the supplementary materials.

\section{Discussion and Conclusions}
In this study, we introduce SurgFusion-Net, a novel adaptive multimodal fusion network for RAS skill assessment, which represents the first validation on clinical cases of vaginal vault closure and dorsal venous complex ligation.
Our model integrates DRA, an innovative attention mechanism featuring adaptive mixed attention and diversified multi-head attention. We present RAH-skill and RARP-skill, two clinical datasets for multimodal skill assessment that capture complex surgical maneuvers and scene transitions in RAH and RARP suturing tasks. The datasets include RGB videos, optical flow, and tool segmentation masks, significantly expanding the scope of AQA in real-world surgical scenarios.

Experimental results on JIGSAWS, RAH-skill and RARP-skill datasets demonstrate SurgFusion-Net's superior performance over the SOTA, achieving significant improvements in both SCC and MAE. Unlike unimodal RGB-based methods that perform well on simulation datasets but struggle to generalize in clinical videos, SurgFusion-Net demonstrates consistent performance across both simulated and clinical environments. Our model is particularly effective in clinical settings with frequent camera movements and background noise where unimodal approaches are compromised.
Ablation studies and qualitative analysis illustrate SurgFusion-Net's effectiveness in adaptively fusing multimodal information through CSFB, DFB and DRA, demonstrating their contribution in handling diverse surgical scenarios.

These findings highlight the potential of SurgFusion-Net as a reliable and interpretable AI assistant for surgical education and clinical skill assessment, offering junior surgeons comprehensive and accurate evaluations and supporting experienced ones in maintaining optimal performance. Future research directions include expanding the datasets to incorporate non-visual cues (e.g., audio, text and kinematic data), integrating segmentation masks for target tissues, and exploring more advanced multimodal fusion mechanisms to further enhance system robustness and usability in clinical settings.

\section{Acknowledgments}
We declare the following conflicts of interest: Danail Stoyanov reports employment with Medtronic Ltd; board membership, consulting or advisory, and equity or stocks with Odin Medical Ltd and Panda Surgical Ltd; and board membership and consulting or advisory with EnActiv Ltd. The remaining authors declare no conflict of interest.

\bibliographystyle{IEEEtran}
\bibliography{mybib}

\end{document}